\documentclass[letterpaper, 10 pt, conference]{ieeeconf}  

\IEEEoverridecommandlockouts                              

\usepackage{chemformula}
\usepackage{ragged2e}
\usepackage{url}

\usepackage{booktabs} 
\usepackage{siunitx} 
\usepackage{caption} 

\usepackage{multirow}
\usepackage{array}
\usepackage{graphicx}

\usepackage{svg}

\usepackage{amsmath}
\usepackage{bm}

\title{\LARGE \bf
State Estimation for Compliant and Morphologically Adaptive Robots
}

\author{Valentin Yuryev$^{1*}$, Max Polzin$^{1}$ and Josie Hughes$^{1}$
\thanks{$^{1}$These authors are with with CREATE Lab, Swiss Federal Institute of Technology Lausanne (EPFL), Lausanne,
Switzerland.}%
\thanks{*Correspondence to: V.Yuryev ({\tt\small valentin.yuryev@epfl.ch})}
}

\begin{document}

\maketitle
\thispagestyle{empty}
\pagestyle{empty}

\begin{abstract}

Locomotion robots with active or passive compliance can show robustness to uncertain scenarios, which can be promising for agricultural, research and environmental industries. 
However, state estimation for these robots is challenging due to the lack of rigid-body assumptions and kinematic changes from morphing. 
We propose a method to estimate typical rigid-body states alongside compliance-related states, such as soft robot shape in different morphologies and locomotion modes. 
Our neural network-based state estimator uses a history of states and a mechanism to directly influence unreliable sensors. 
We test our framework on the GOAT platform, a robot capable of passive compliance and active morphing for extreme outdoor terrain. 
The network is trained on motion capture data in a novel compliance-centric frame that accounts for morphing-related states. 
Our method predicts shape-related measurements within 4.2\% of the robot's size, velocities within 6.3\% and 2.4\% of the top linear and angular speeds, respectively, and orientation within 1.5$\unit{\degree}$. 
We also demonstrate a 300\% increase in travel range during a motor malfunction when using our estimator for closed-loop autonomous outdoor operation.

\end{abstract}

\section{Introduction}
\label{sec:intro}

\begin{figure*}[!tbp]
    \centering
    \includegraphics[width=0.95\textwidth]{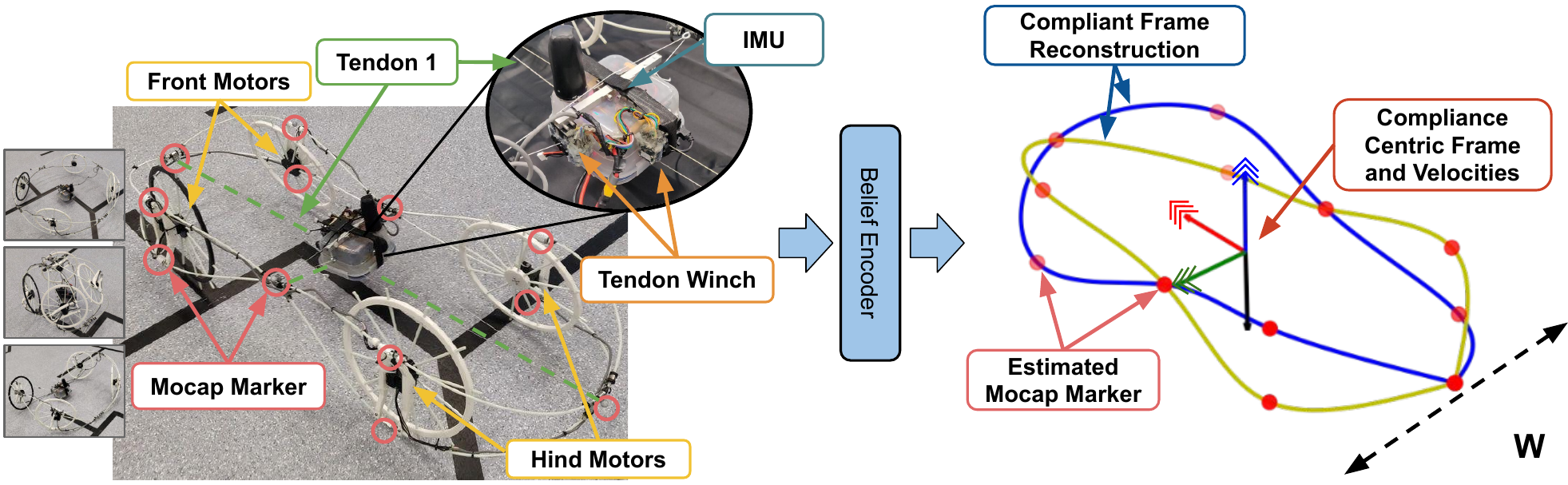}
    \caption{Overview of our state estimation framework for compliant and morphologically adaptive platforms. Left: placement of motors, sensors and motion capture for data collection. Right: reconstruction of compliance-centric frame, shape and width.}
    \label{fig:markers}
    \vspace{-4mm}
\end{figure*}

For robots to support outdoor industries and research activities, such as animal monitoring, climate surveillance, and agriculture, they require the capability to operate within and traverse extreme terrain conditions \cite{angeliniRoboticMonitoringHabitats2023}\cite{dunbabinRobotsEnvironmentalMonitoring2012}\cite{kingTechnologyFutureAgriculture2017}. 
While animals demonstrate such capabilities, and can operate on challenging and varied terrains, typical state-of-the art robotic solutions, such as rigid-body quadrupeds, are primarily confined to challenging but man-made terrain \cite{ramdyaNeuromechanicsAnimalLocomotion2023}.  
This can be partially attributed to their reliance on a singular mode of locomotion and an inability to physically adapt to the wide range of conditions found in outdoor environments \cite{leeLearningQuadrupedalLocomotion2020}\cite{MikiLearningRobustPerceptive2022}.
Conversely, locomoting animals exploit passive and active compliance to achieve adaptability and beneficial interaction with the environment. 
For example passive adaptation enables tumbleweeds to roll in the wind, cockroaches to squeeze through tight spaces, and kangaroos to store energy in their legs\cite{jayaramCockroachesTraverseCrevices2016}\cite{morganMeasurementsMuscleStiffness1978}. 
Active morphological adaptability is exemplified in nature by armadillos or hedgehogs, which can modify their shape to defend themselves or leverage terrain to their advantage \cite{stappDefensiveDiversityFactors2022}\cite{bergerUrbanHedgehogBehavioural2020}.

This biological inspiration has inspired a paradigm shift in robot design toward systems that leverage passive compliance and adaptive morphologies\cite{stellaScienceSoftRobot2023a}, resulting in a new robot design space for traversing varied terrain \cite{wilcoxRapidReversibleMorphing2025}\cite{kimBioinspiredDualmorphingStretchable2019}. 
To fully exploit the autonomy of such robots, reliable motion controllers and high-level decision-making modules are required, which are dependent on reliable state estimation. 
For compliant, morphing robots this becomes challenging due to kinematic changes from active morphology and the lack of rigid-body assumptions.
Our goal is to produce a reliable state estimator that accommodates the challenges from robots with passive compliance and active morphological adaptability.
For this paper, we focus our efforts on the GOAT robot, a elastic-rod shelled robot capable of driving, rolling and swimming by actively changing its morphology \cite{PolzinRoboticLocomotionActive2025}.


State estimation for rigid-body robots is a widely researched topic \cite{kalmanNewApproachLinear1960}\cite{barfootStateEstimationRobotics2024}. 
For legged robots, modern actuation and sensor advancements have enabled precise model-based state estimation \cite{bloeschStateEstimationLegged2013}. 
Similarly, in autonomous driving, deep-learning methods have provided reliable state estimation in various dynamic environments \cite{ghoraiStateEstimationMotion2022}.
While robust and industry-proven, these methods rely on the rigid-body assumption. 
For compliant, soft robots, the information encoded in the relationship between the robot-centric frame and the rest of the robot's kinematics—such as the location of the sensors relative to the motors—becomes inconsistent and unobservable.
The lack of a rigid frame further complicates state estimation as the computational and sensor payloads can experience large impacts and perturbations.

With the increasingly prevalence of soft robotics, there have been state-estimation methods developed for compliant systems both on the algorithm side such as kinematic-based filtering \cite{stellaSoftRobotShape2024} and sensor side such as IMU arrays and piezoelectric sensors \cite{woodmanStretchableShapeSensing2025}.  
However, these primarily focus on minor deformations, not significant morphological change during operation. 
Recent AI advances have produced learning-based approaches that accommodate active morphological variations within a single locomotion mode, such as walking, but not for robots capable of multiple modes like driving and rolling \cite{luoMorALLearningMorphologically2024}.
To the best of our knowledge, no state estimation method exists that combines different locomotion modalities with robot compliance.



We propose a state estimation approach for compliant robots capable of active morphological adaptation, shown on Fig. \ref{fig:markers}. 
Our method utilizes a machine learning-based history belief encoder, trained on motion capture data, to approximate a robot's typical rigid-body states (e.g., linear and angular velocities) and states unique to compliant and morphologically adaptive robots (e.g. outer shell shape).
To accommodate passive compliance, we introduce a compliance-centric frame based on the overall robot shape, rather than a classical robot-centric frame tied to sensor positions. 
To support various locomotion modalities, we augment a feedforward PID controller with soft-body-specific information from the state and shape estimator.


We test our method on the GOAT platform, which features passive terrain compliance and active morphological changes for different locomotion modalities like driving and rolling \cite{PolzinRoboticLocomotionActive2025}. 
We demonstrate state estimation for both rigid-body states (e.g., linear and angular velocities) and morphing-robot-specific states (e.g., outer shell shape) across varied morphological scenarios. 
Model performance is validated against motion capture ground truth.
We showcase improved control via a morphologically adaptive PID controller that enhances velocity tracking during driving. 
To demonstrate robustness, we test the history belief encoder with malfunctioning morphology sensors in unforeseen corner cases. 
Finally, we stress-test our closed-loop controller, which relies on our state estimator, in an outdoor scenario against an open-loop baseline.

The remainder of this paper is structured as follows. 
First, we introduce our method.
We then present the experimental setup and results. 
Finally, we summarize our findings and propose future research directions.

\section{Methods}
\label{sec:methods}
This section details the methodology for frame and state estimation of a compliant, morphologically adaptive robot. 
We introduce the compliance-centric frame, our state estimation framework, data collection procedures, noise resistance, and the feedforward PID morphing controller that utilizes our estimator.



\subsubsection{Compliant, Morphing Robot Platform}

We develop and deploy our method on the GOAT robot \cite{PolzinRoboticLocomotionActive2025}, a platform designed for extreme outdoor traversal, shown in Fig. \ref{fig:markers}.
The robot features a compliant elastic rod frame with four direct drive motors.
Two winches and tendons enable the frame to be morphed and folded.
By shortening or elongating the tendons, GOAT modulates its forward mobility and turning ability, and can morph into a ball to roll—an effective method for traversing steep terrain.
The robot operates on grassy, rocky, and sloped terrains, and can swim. Its payload, which contains the compute and sensors, is suspended within the compliant frame via the tendons.

Autonomous control requires reliable state estimation to determine the robot’s velocities and orientation.
Shape estimation is also critical to achieve target morphologies reliably, as relying solely on tendon length is error-prone due to encoder inaccuracies, tendon slippage, or malfunction.
We propose estimating the state and shape using only onboard sensing: an IMU in the central payload, and motor data from the wheels (current, measured velocities, and commanded velocities) and winches (estimated tendon length). 
Although demonstrated on the GOAT platform, our state-estimation methodology generalizes to other compliant or morphing robots.

\subsubsection{Compliance-Centric Frame}
\label{sec:compliance_centric_frame}

Compliance and adaptive morphology challenge velocity and configuration estimation due to fundamental kinematic changes and significant base frame oscillation. 
Typically, state estimation for rigid robots uses a body- or sensor-centric reference frame. 
For compliant, morphing robots, we propose a compliance-centric frame whose position and orientation depend on key points along the compliant structure.

For GOAT, this frame is constructed by defining a unit X-vector from the "front" and "back" points (Fig. \ref{fig:markers}). 
A unit Y-vector is similarly defined from the "left" and "right" points. 
The Gram-Schmidt method ensures orthogonality between the X and Y components. 
The Z-vector is the cross product of X and Y. 
This approach minimizes oscillation noise and simplifies estimator training.


\subsection{RNN and Belief Encoder}

\begin{figure}[tb]
    \centering
    \includegraphics[width=\linewidth]{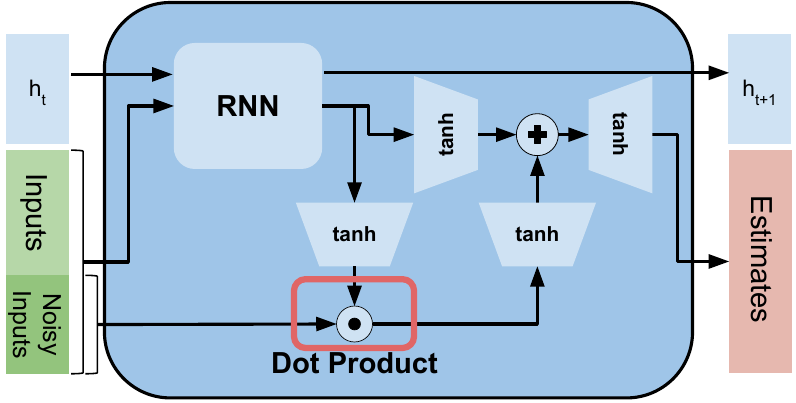}
    \vspace{-4mm}
    \caption{Belief encoder architecture. Hidden states are passed through onto the next iteration. Inputs are fully fed into the RNN while a subset of the inputs that are considered untrustworthy are further adjusted by the network.}
    \label{fig:network}
    \vspace{-4mm}
\end{figure}

Our main contribution is a state and shape estimator for compliant, morphologically adaptive robots. 
For GOAT, this estimates the compliance-centric frame's velocities, orientation, and key outer frame points from which we can then estimate the frame shape (labeled as motion capture markers in Fig. \ref{fig:markers}) using only an IMU, motor and winch measurements.


To address changing morphology, robot compliance and sensor uncertainties we employ a history-based belief encoder (Fig. \ref{fig:network}), which has proven effective for estimating terrain from noisy perception data \cite{MikiLearningRobustPerceptive2022}. 
This network compensates for unreliable measurements by incorporating historical data via a recurrent neural network (RNN). 
The RNN output can directly interact with untrustworthy sensor inputs, enabling real-time error correction. 
We adapt this approach to handle generally unreliable sensor data.

In our implementation, we replace the self-attention layer with a dot product between estimated and error-corrected components. 
We use hyperbolic tangent activation units, as it is meant to work better with contentious input sensor data that revolves near zero.

\subsection{Data Collection and Training}
To collect ground truth data for training the shape and state estimator we use a motion capture system. 
We manually commanded the robot to drive, turn, and morph, collecting data of it driving, morphing, and rolling within a motion capture system spanning approximately 3×4 m. 
To ensure a balanced dataset, we included an even mixture of static cases, where the robot held various shapes without commanded motion, and dynamic cases, where it drove forward or turned at various speeds.

To estimate the shape of GOAT's outer compliant frame, we estimate frame points along the two elastic rods, matching the motion capture marker locations in Fig. \ref{fig:markers}. 
We transform the world centric motion capture to a compliance-centric frame via a homogeneous transformation matrix to provide the target training data for our belief encoder: the frame point locations (X, Y, Z) in the compliance-centric frame.
To reconstruct and visualize the 3D shape, we fit two cubic splines (one per elastic rod) to the estimated frame points, as shown in Fig. \ref{fig:markers}.

For velocity estimation, ground truth velocities are computed in the compliance-centric frame using the finite difference method, and an exponentially moving average (EMA) filter applied to reduce the noise. 
The belief encoder is trained to estimate these smoothed velocities.
The network's final output is the gravity vector expressed in the compliance-centric frame.

The network is trained via supervised learning to minimize the RMSE between ground truth and predicted values, as shown in Eq. \eqref{eq:main_objective}. 
Due to a large quantity of near-zero outputs, we augmented the cost function during training in a style similar to mean absolute percentage error (MAPE), as shown in Eq. \eqref{eq:main_objective}, giving more weight to values near zero \cite{botchkarevPerformanceMetricsError2019}.

\begin{equation}
\mathcal{L}(\bm{\theta}) = \min_{\bm{\theta}} \sum_{i=0}^{N} \frac{\left( f(\bm{x}_{i}; \bm{\theta}) - y_{i} \right)^2}{1 + |y_{i}|}
\label{eq:main_objective}
\end{equation}

\noindent where $\bm{x}$ is the model's inputs, $\bm{\theta}$ is the model's parameters, $f(\bm{x}; \bm{\theta})$ is the model's prediction function and $y$ is the target value from the training data.

\subsection{Robustness via Noise Addition}
During outdoor operation, the compliant frame often leads to significant oscillations and sporadic noise in sensor readings. 
To ensure robustness to these perturbations, we augment our training data by adding Gaussian and sinusoidal noise. 
Furthermore, sensor alignment and controllability (e.g., the tendon connection between the frame and payload) can be compromised. 
To ensure robustness to such cases, we also augment the data with time-consistent measurement offsets.

\subsection{Feedforward PID Controller for Morphologically Adaptive Robot}
To exploit the state-estimator, we implemented a feedforward velocity-tracking PID controller (Eq. \eqref{eq:discrete_pid}) for the forward linear and yaw angular components. 
The integral term of the PID includes a wind-down coefficient $\alpha < 1$ to prevent saturation:

\begin{equation}
e_{k} = \hat{v}_{k} - v_{k}
\label{eq:velocity_error}
\end{equation}

\begin{equation}
\hat{u}_{k} = \hat{v}_{k} + K_{p} e_{k} + K_{i} \sum_{i=0}^{k} e_{i} * \alpha^{k-i} + K_{d} (e_{k} - e_{k-1})
\label{eq:discrete_pid}
\end{equation}


\noindent where $e_{k}$ is the error at control timestep $k$, $\hat{u}_{k}$ is the desired velocity of the PID controller $\hat{v}$ and  $v$ are desired and measured velocities, $K$ are the PID gains, $k$ is the control timestep.

We then utilize the linear forward component $\hat{u}_{x}$ and angular yaw $\hat{u}_{\phi}$ component of the PID Eq. \ref{eq:discrete_pid} output $\hat{u}$ to compute the motor velocities $v_{\text{left}}$ and $v_{\text{right}}$:

\begin{equation}
v_{\text{left}} = \hat{u}_{x} - \frac{W}{2} \hat{u}_{\phi} \qquad
v_{\text{right}} = \hat{u}_{x} + \frac{W}{2} \hat{u}_{\phi}
\label{eq:wheel_velocities}
\end{equation}

However, a morphologically adaptive robot such as the GOAT does not have a constant width $W$, shown in Fig \ref{fig:markers}, as the robot can change and morph its shape during operation.
Thus, we use our shape estimation framework to estimate $W$ as the Euclidean distance between the averaged points $\bm{p}_{i}$, as shown in Eq. \ref{eq:width}, attached to the motors in Fig. \ref{fig:markers}:

\begin{equation}
W = \| \bm{p}_{\text{left}} - \bm{p}_{\text{right}} \|_2 = \| \frac{1}{N} (\sum_{i=1}^{N} \bm{p}_{\text{left},i} - \sum_{i=1}^{N} \bm{p}_{\text{right},i}) \|_2
\label{eq:width}
\end{equation}

This dynamic assessment of $W$ is then used in Eq. \ref{eq:wheel_velocities} to determine the appropriate wheel velocity.

\section{Experimental Setup}
\label{sec:experimental_setup}

Our state estimation framework for compliant robots is applied to a modified version of the GOAT robot shown in Fig. \ref{fig:markers} \cite{PolzinRoboticLocomotionActive2025}. 
We retrofitted the robot with Dynamixel XM430-W350 motors to collect reliable proprioceptive data, including motor velocities and currents. 
Additional sensor data for state-estimation comes from an IMU mounted in the central, tendon suspended payload which reports linear accelerations, angular velocities, and orientation estimates. 
As a potentially noisy input, as shown in Fig. \ref{fig:network}, we use the lengths of the tendons connecting the robot's payload to its compliant morphing outer frame, which is estimated from the open loop winch motors encoders.
To collect ground truth training data, we use motion capture system OptiTrack PrimeX22 cameras with sub-millimeter accuracy \cite{krumpekReproducibleOpticalTracking2025}.
To sync the data from the GOAT platform and the motion capture system, we do a post-processing run, where we manually match the change in marker positions to a timestep where a command is given to the robot, such as to drive forward or to morph.

For training, we utilize a mean absolute percentage error loss objective with truncated backpropagation through time for 50 timesteps. All inputs and outputs are normalized. We train our model using a PyTorch-based framework on an RTX 5090 GPU. The total amount of data trained on is approximately 40 minutes of operation at 20Hz. Each data set was separately augmented with gaussian noise, constant offsets and sinusoids, resulting in 4 times the data amounting to 160 minutes. The latent space and RNN hidden dimension are both set to 32. The network input dimension is 21, consisting of: the IMU gravity vector (3), IMU linear acceleration (3), IMU angular velocity (3), left and right wheel motor velocity commands (2), measured motor currents (4), and measured motor velocities (4). The output dimension is 45, corresponding to: 12 frame points (each 3D), linear velocities (3), angular velocities (3), and the gravity vector (3) in the wheel-centric frame.

For computation onboard the robot, we use a Raspberry Pi 5 (8GB) running a PyTorch JIT-traced policy. The system processes inputs and outputs at 20 Hz. All inputs and outputs are normalized and denormalized at this frequency to match the training conditions. 

For linear velocity tracking experiments, we utilized P, I, and D gains of 1, 0.0, and 0.1, respectively.
We tracked the desired yaw rate using P, I, and D gains of 1, 0.01, and 0.01, respectively. 
All gains were tuned empirically on the robot. 

All reported results are a mixture of offline (computed from a recording) and online (computed on the robot) data in no way used during training, representing unbiased estimator's performance in a real-world scenario.

\section{Experimental Results}
\label{sec:results}

\begin{table*}[!t]
    \centering
    \caption{RMSE of reconstructed robot frame points, gravity vector and velocity components across different scenarios}
    \label{tab:recon_rmse}
    \begin{tabular}{@{}lccccccccccr@{}}
    \toprule
     
    \multirow{2}{*}{\textbf{Scenario}} & \multirow{2}{*}{\textbf{Frame Points (\unit{\mm})}} & \multirow{2}{*}{\textbf{Gravity (\unit{\degree})}} & \multicolumn{4}{c}{\textbf{Linear Velocity (\unit{\mm\per\second})}} & \multicolumn{4}{c}{\textbf{Angular Velocity (\unit{\degree\per\second})}} \\
    \cmidrule(lr){4-7} \cmidrule(lr){8-11}
     & & & $u_x$ & $u_y$ & $u_z$ & \textbf{Mean} & $u_{\psi}$ & $u_{\theta}$ & $u_{\phi}$ & \textbf{Mean} \\
    \midrule

(A) Static Circle & 26 & 1.1 & 3 & 1 & 2 & 2 & 0.4 & 0.1 & 0.2 & 0.2 \\
(B) Static Rover & 44 & 2.3 & 4 & 7 & 2 & 5 & 0.2 & 0.2 & 0.6 & 0.4 \\
(C) Static Sphere & 50 & 1.1 & 2 & 11 & 3 & 5 & 0.6 & 0.3 & 0.6 & 0.5 \\
(D) Static Intermediate w/ Tendon & 53 & 1.4 & 14 & 6 & 4 & 8 & 0.6 & 0.5 & 1.0 & 0.7 \\
(E) Static Intermediate w/o Tendon & 76 & 2.1 & 8 & 6 & 1 & 5 & 0.3 & 0.2 & 0.5 & 0.4 \\
Driving Forwards & 28 & 1.7 & 26 & 30 & 4 & 20 & 1.1 & 0.8 & 1.6 & 1.1 \\
Yawing in Place & 40 & 1.3 & 16 & 11 & 6 & 11 & 1.0 & 1.0 & 4.4 & 2.2 \\
Driving Forwards w/ dist. & 16 & 0.8 & 33 & 16 & 6 & 18 & 1.4 & 1.1 & 2.4 & 1.7 \\
\midrule
\textbf{Mean} & 42 & 1.5 & 13 & 11 & 4 &  & 0.7 & 0.5 & 1.4 &  \\
\textbf{Std Dev.} & 18 & 0.5 & 10 & 8 & 2 &  & 0.4 & 0.4 & 1.3 &  \\
    \bottomrule
\end{tabular}
\vspace{-4mm}
\begin{minipage}{\columnwidth}
\end{minipage}
\end{table*}

In this section, we first report the performance of the state estimation, before demonstrating the role of this state estimation for real-time control. 

\subsection{State Estimation}

To obtain a quantitative assessment of the state estimation, the root mean square error (RMSE) between the estimation and motion capture ground truth was determined. This includes the compliance-centric frame points (12 motion capture points), angle between true gravity and estimated gravity, linear and angular velocities, shown for static and dynamic scenarios. The results are shown in Table \ref{tab:recon_rmse}. Static scenarios indicate cases where the robot does not move and are primarily used to validate the shape reconstruction. Dynamic cases, such as robot driving or turning, are when there is a desired velocity command given to analyze gravity and velocity estimation.

\subsubsection{Static Shape Reconstruction}

\begin{figure}[tb]
    \centering
    \includegraphics[width=\linewidth]{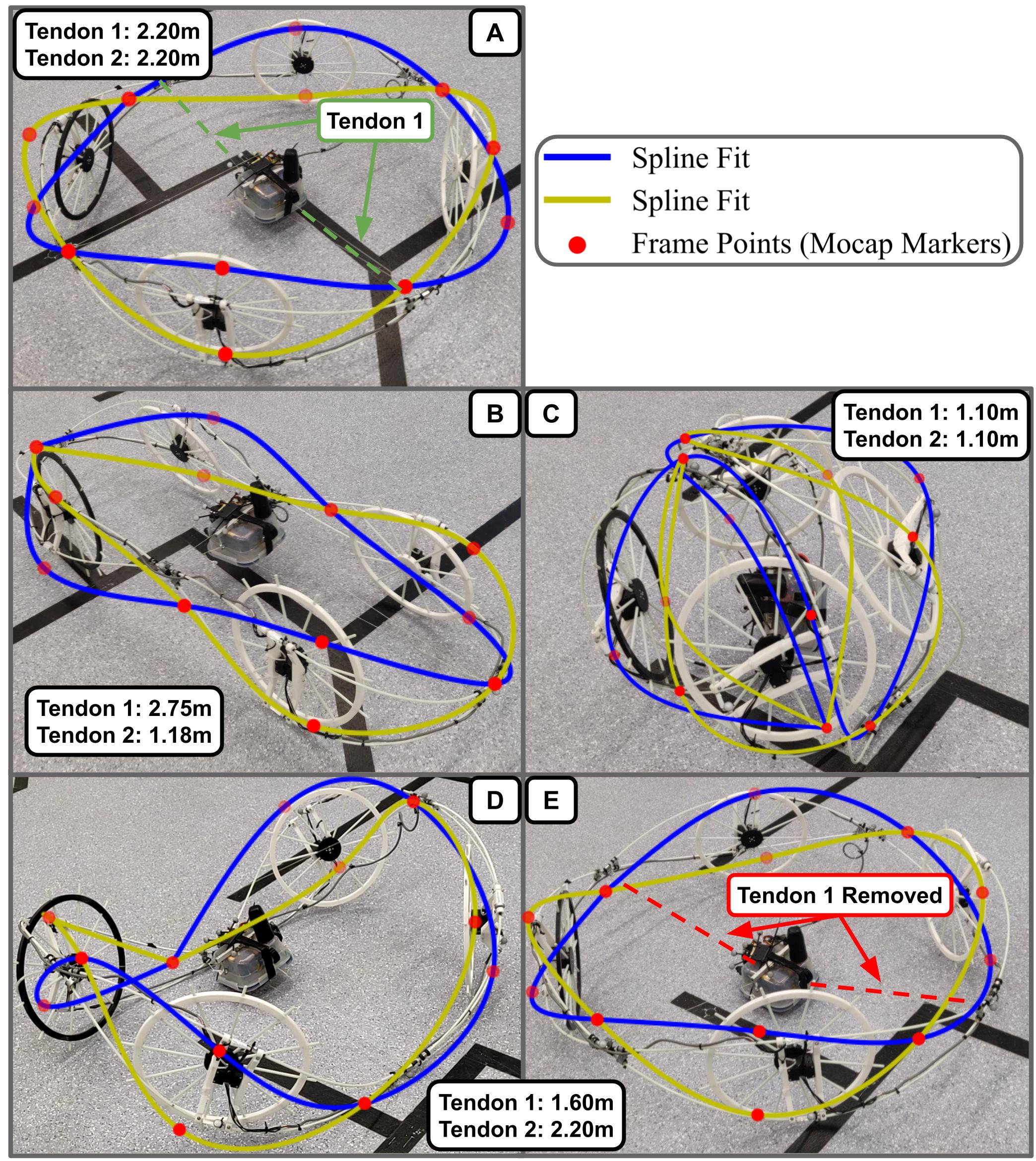}
    \vspace{-4mm}
    \caption{Overlay of the reconstructed frame shape over the actual robot image for scenarios in Table \ref{tab:recon_rmse}. Note that tendon lengths estimated specified in case (D) and (E) are the same. Yet the framwork is capable of reconstruction the shape.}
    \label{fig:reconstruction}
    \vspace{-4mm}
\end{figure}

To evaluate shape reconstruction performance, we selected various static scenarios where the robot maintains a fixed shape without moving or morphing. The selected shapes include a circle (A), rover (B), sphere (C), and intermediate pose (D, E), depicted in Fig. \ref{fig:reconstruction}. Cases (A-C) represent the main operating modes of the GOAT robot, namely turning, driving, and rolling. For visualizing the estimated shape, we apply a twice differentiable cubic spline to the estimated points. The RMSE is calculated from 5 seconds of time series data at 20Hz. For the standard configurations in \ref{fig:reconstruction} (A-C), the RMSE error is given in Table \ref{tab:recon_rmse}. The mean point reconstruction of the main operating modes (A - C) are in 2 to 5 cm range, which from visual qualitative assessment in Fig. \ref{fig:reconstruction} provides a good visual reconstruction of the shape of the robot. Over all scenarios, the error remains approximately 4.2\% of the overall span of the robot that has a diameter of 1 meter when in circular form.

To evaluate the performance of the shape estimation in scenarios where there is error or disturbance in the tendon measurements or configuration, we simulate a scenario where one tendon controlling the morphology is removed, as shown in Fig. \ref{fig:reconstruction} scenario (E). 
We then compare the resulting estimated shape against case (D), where the tendon estimated measurements is the same but the tendon is still attached, resulting in frame deformation and the saddle shape. 
The overall shape remains circular and closer to the real circular shape of the robot (A), rather than potential intermediate shape (D), that the network would estimate if it only relied on tendon measurements. Frame point reconstruction RMSE for that scenario is 76$\unit{\mm}$, which is higher than the main operating modes (A - C) but is still visually representative of the robot's shape.
This indicates that our belief encoder adjusts its estimates and discards the inaccurate tendon length measurements provided to it, showcasing the belief encoder-based network capability to adapt to corner-case situations where certain sensor measurements are untrustworthy. 

\subsubsection{Dynamic Shape Reconstruction}

\begin{figure*}[tbp]
    \centering
    \includegraphics[width=0.95\textwidth]{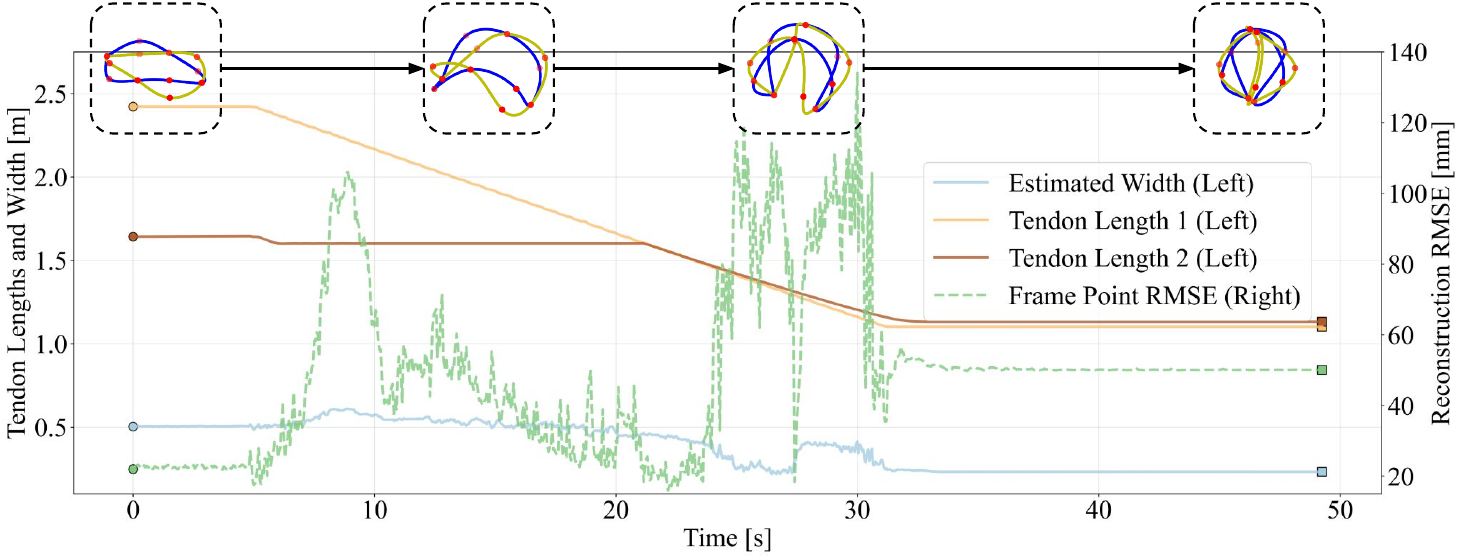}
    \vspace{-4mm}
    \caption{Time series of robot morphing from rover (scenario B) to sphere (scenario C) formation.}
    \label{fig:morph_timeseries}
    \vspace{-4mm}
\end{figure*}

The shape estimation can also be used "live" throughout the robot's morphing process. Fig. \ref{fig:morph_timeseries} shows the time series response of the robot morphing from a flat spherical scenario (C) to a circular scenario (A) (Fig. \ref{fig:reconstruction}), estimated width $W$ of the robot and the RMSE of frame point estimation throughout the process. We observe that the static shapes are easier to predict, while the RMSE of the shape estimator increases during the intermediate morphing phases. This may be explained by the belief encoder not fully trusting the estimated tendon lengths until the IMU information has provided more confirming data, or due to lack of training data in these specific scenarios as they are very transient.

The estimation of the robots width $W$, also shown in Fig.\ref{fig:morph_timeseries}, is directly reliant on the performance of our frame point reconstruction and influences the PID controller. While the intermediate shape error results in incorrect calculation of the width, the final stable configuration RMSE, such as the rover, is low enough that we can reliably use its estimates for adjusting our low level wheel controller described in Eq. \eqref{eq:wheel_velocities}. 

\subsubsection{Gravity}

\begin{figure}[tb]
    \centering
    \includegraphics[width=\linewidth]{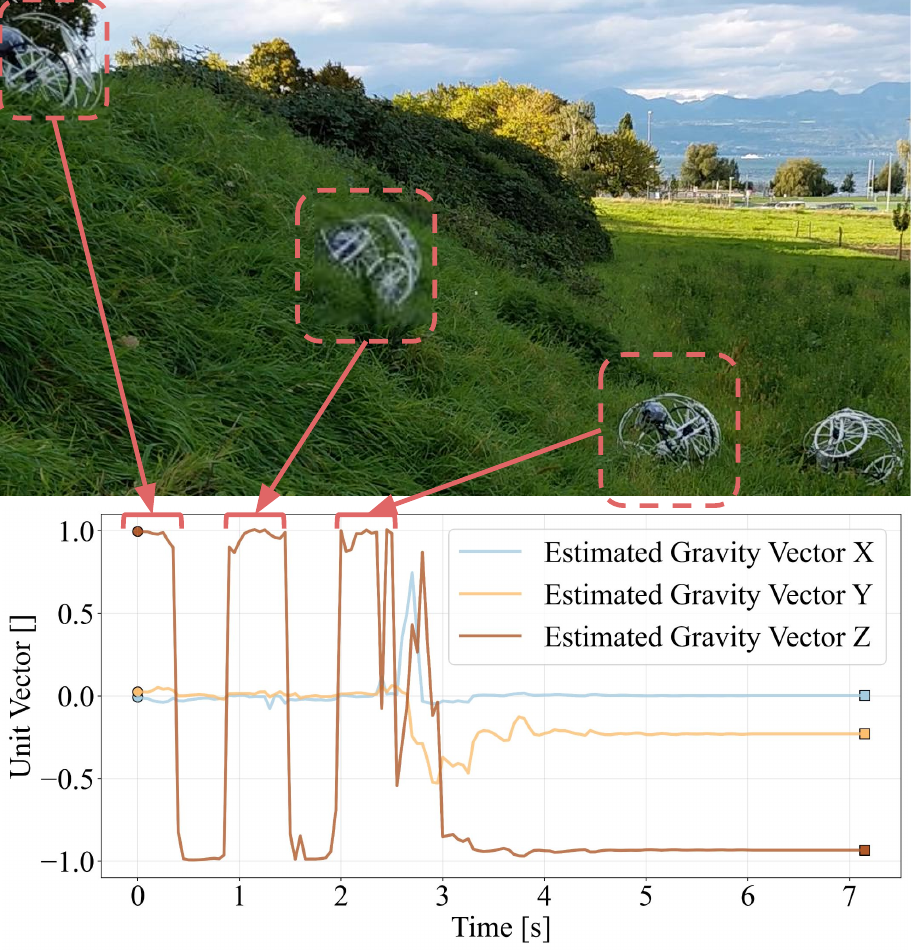}
    \vspace{-4mm}
    \caption{(Top) Time lapse images of the GOAT robot rolling down a hill and (Bottom) the estimated gravity vector time series.}
    \label{fig:gravity}
    \vspace{-4mm}
\end{figure}

While gravity estimation is not directly used by our controller, it is a valuable component of state estimation in scenarios where the robot changes its shape and rolls. Note that IMU linear acceleration estimates this value almost directly. As such, we expect the measurement to be accurate in static cases, but not exact since we estimate the gravity frame in our compliance-centric frame as we defined in Section \ref{sec:compliance_centric_frame}. All gravity error values are below 3$\unit{\degree}$, representing adequately accurate gravity vector estimation for estimating robot orientation.

To demonstrate how our method would perform in an outdoor rolling locomotion mode, we have also performed a short rolling of a hill experiment to demonstrate real-world estimation of the gravity vector in a dynamic cases, which are harder to read from the linear accelerations of the IMU. Fig. \ref{fig:gravity} qualitatively shows, that we can approximate the number of rolls and orientation whilst in a sphere shape rolling down a hill. Note that, while the estimator was trained on normalized gravity vector, nothing is enforcing this behavior for the output of our estimator.

\subsubsection{Linear \& Angular Velocities}

While we estimate all linear and angular velocities, the relevant components for operating the GOAT robot are the linear forward velocity ($u_{x}$) and the angular yaw rate ($u_{\phi}$). 
Therefore, we focus our analysis primarily on these components. The estimation results for all other components across different scenarios can be found in Table \ref{tab:recon_rmse}. 

For the static cases (A - C), we collected 5 seconds worth of data where the robot does not receive any desired velocity or morphing commands and does not move. In these scenarios the linear and angular velocity error is significantly lower than dynamic cases since the estimates are all close to 0, the easiest case to predict. Since the network was trained on some rolling and non static data, the estimates are never exactly zero, resulting in minor estimation error.

\begin{figure}[tb]
    \centering
    \includegraphics[width=\linewidth]{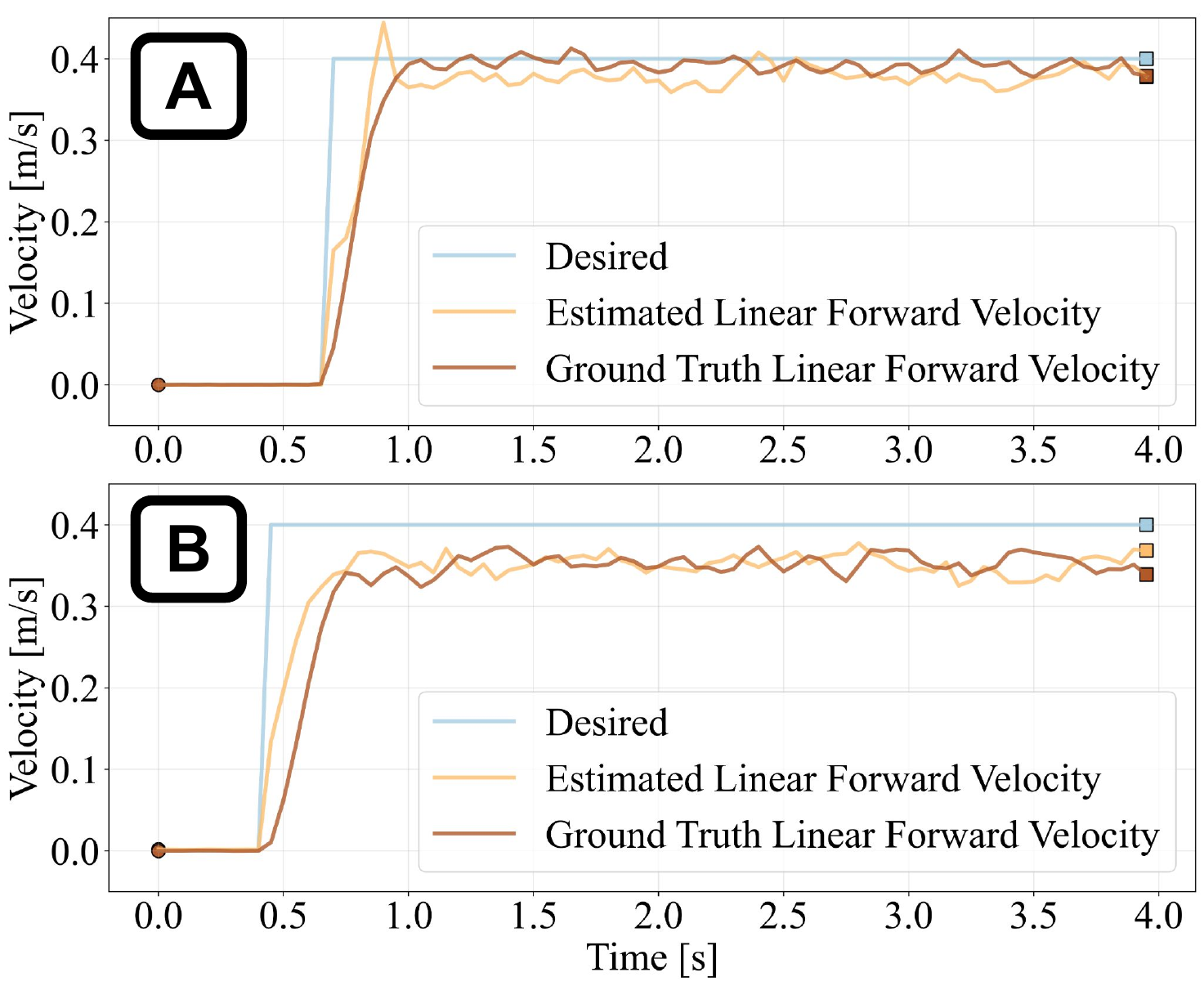}
    \vspace{-4mm}
    \caption{Estimation of linear tracking while driving forwards with closed loop controller with disturbance (A) and with simulated disturbance of a broken wheel (B).}
    \label{fig:linear_tracking}
    \vspace{-4mm}
\end{figure}

In dynamic scenarios, we provide a linear or angular velocity command to the robot and observe the step responses. The compliance-centric frame is computed from the ground truth motion capture data, which is then used to calculate ground truth velocities via the finite difference method. We expect the dynamic cases to be harder to predict due to noise from the difficulty of integrating sensor data (IMU only provides linear accelerations), sensor payload disturbances (oscillation due to compliance) while moving and wheels slipping. To evaluate our linear forward $u_x$ velocity estimation, we executed a pure forward motion command within the motion capture facility with and without a simulated disturbance of a broken wheel, shown in Fig. \ref{fig:linear_tracking}. The error for driving forwards is 26$\unit{\mm}$ for nominal scenario (Fig. \ref{fig:linear_tracking}.A) and 33$\unit{\mm}$ for the scenario with a simulated "broken" wheel (Fig. \ref{fig:linear_tracking}.B). Given maximum commanded velocity of 400$\unit{\mm\per\second}$, this results in error of 6.5\% and 8.3\%, respectively.

\begin{figure}[tb]
    \centering
    \includegraphics[width=\linewidth]{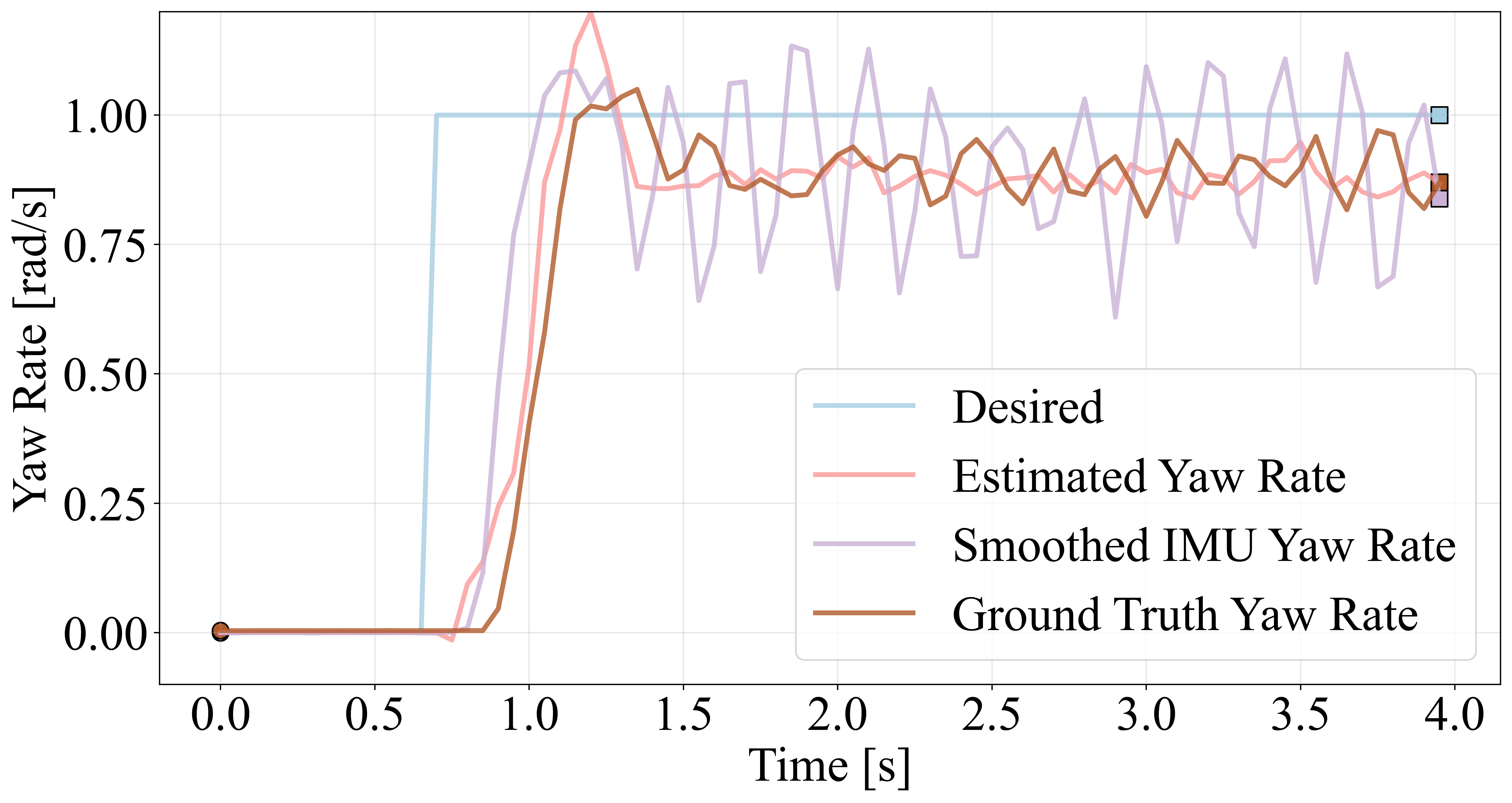}
    \vspace{-4mm}
    \caption{Estimation of yawing rate during pure yawing command tracked with our PID controller. IMU data is based on IMU installed inside the robot payload and does not fully correspond to the yaw rate that is estimated.}
    \label{fig:yaw_estimation}
    \vspace{-4mm}
\end{figure}

To validate our yaw rate estimates, we performed a pure yawing motion inside the motion capture system for comparison against the smoothed finite-difference velocities derived from the ground truth data. The RMSE of the yaw estimation was calculated to be 4.4$\unit{\degree\per\second}$, as shown in Fig. \ref{fig:yaw_estimation} which is 2.4\% of the maximum commanded yaw rate of 1$\unit{\radian\per\second}$. The resulting estimates are close to the ground truth data and raw IMU data. Note that the figure incorporate IMU readings, which were processed through an exponential moving average (EMA) filter with $\alpha = 0.3$. While the IMU measurements provide a rough quality check, the IMU frame is located inside the robot's payload (as shown in Fig. \ref{fig:markers}) and does not align with our compliance-centric frame, calculated from the points attached to the motors.

\subsection{Morphing Feed Forward PID Control}

In this section, we discuss the validation of our morphing feedforward PID controller and its use of state estimation.

\begin{figure}[tb]
    \centering
    \includegraphics[width=\linewidth]{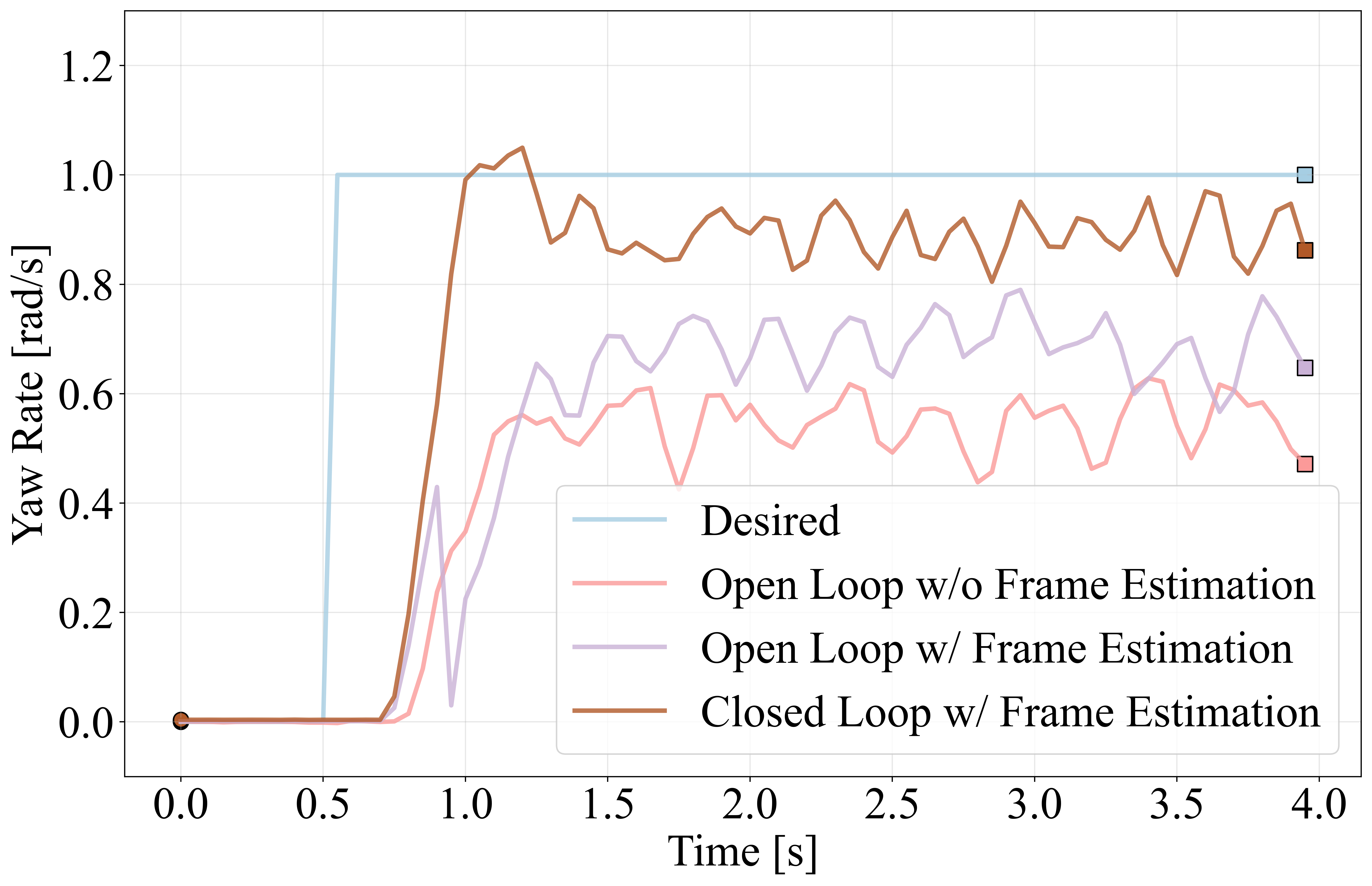}
    \vspace{-4mm}
    \caption{Motion capture based ground truth yaw rate tracking for scenarios with frame estimation disabled, enabled in open and closed loop scenarios.}
    \label{fig:yaw_rate}
    \vspace{-4mm}
\end{figure}

To showcase the benefits of online state-estimation of the compliant frame for robot control, more specifically the dynamic width estimation, we performed a pure yaw rate experiment comparing three configurations: a constant frame width estimated at 0.5$\unit{\m}$ (the width in rover form), dynamic frame estimation without PID control, and the full system with dynamic frame width estimation and the feedforward PID controller. We then compared the ground truth yaw rates from the motion capture system. 
In Fig. \ref{fig:yaw_rate}, it can be seen that the desired yaw rate tracking steady-state error improves with dynamic frame estimation from 0.45$\unit{\radian\per\second}$ to 0.31$\unit{\radian\per\second}$ (31\% improvement) and is further enhanced to 0.11$\unit{\radian\per\second}$ (65\% improvement) with the addition of the PID controller.

\begin{figure}[tb]
    \centering
    \includegraphics[width=\linewidth]{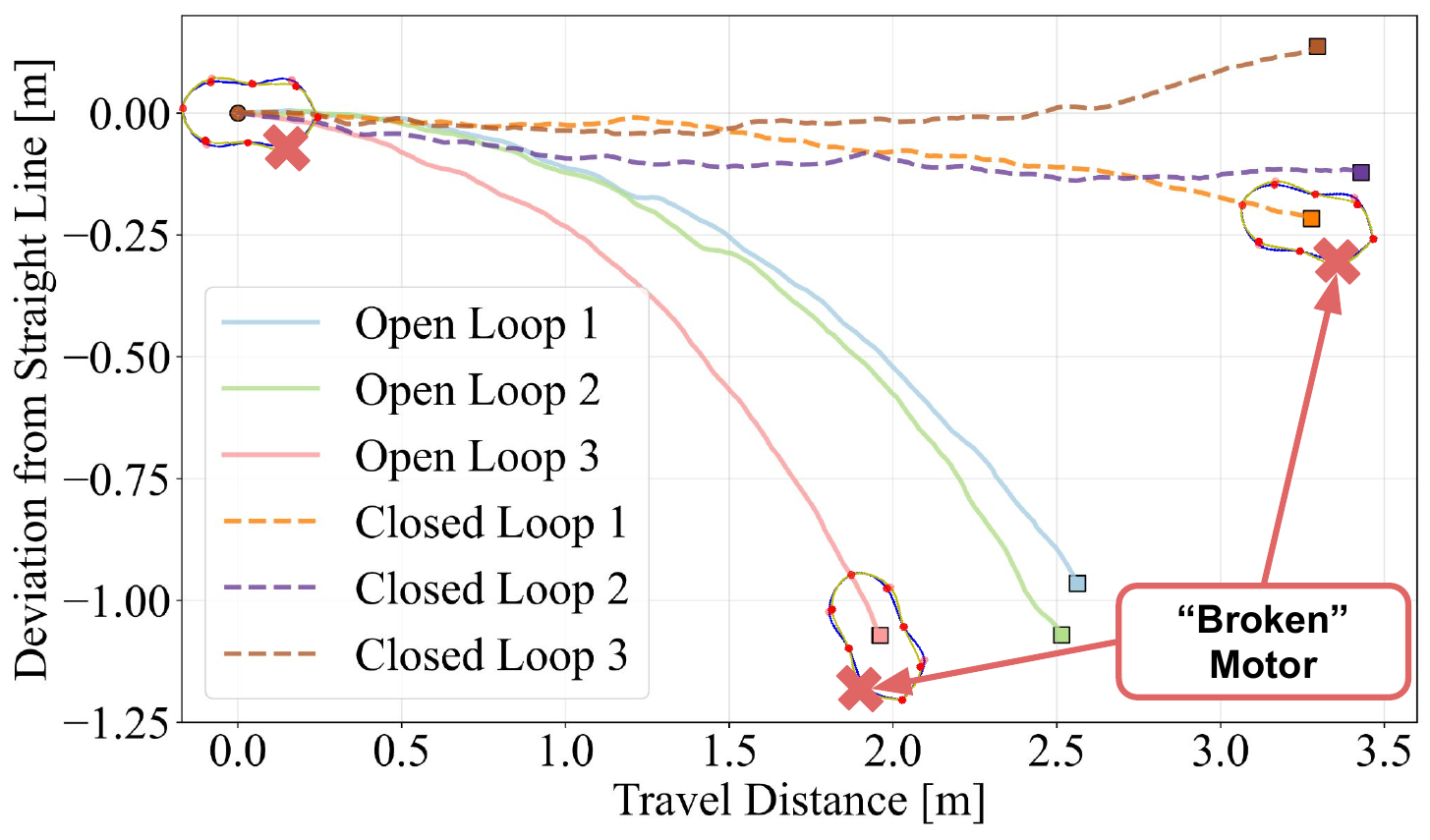}
    \vspace{-4mm}
    \caption{Top-down view of motion capture based open loop vs closed loop forward driving performance when encountering a malfunctioning front-right motor.}
    \label{fig:top_view}
    \vspace{-4mm}
\end{figure}

To test whether our state estimation is sufficient for controlling the robot's desired velocities, we performed a straight-line driving experiment inside the motion capture system. The target forward linear velocity $\hat{u}_x$ was set to a constant 0.4 m/s without the additional PID control from Eq. \eqref{eq:discrete_pid}, while the desired yaw rate was set to 0$\unit{\degree\per\second}$. To further stress-test the framework, we multiplied the commanded velocity of the front-right wheel by a factor of 0.75 to simulate a motor malfunction. Without feedback, the robot's open-loop controller causes it to veer to the right, as shown in Fig. \ref{fig:top_view}. In contrast, the closed-loop controller—relying on our state estimation—compensates for the yaw-rate error and maintains a straighter trajectory. This demonstrates our framework's capability to provide accurate measurements that enable a controller to generate reliable wheel commands.

\label{sec:experimental_setup}
\begin{figure}[tb]
    \centering
    \includegraphics[width=\linewidth]{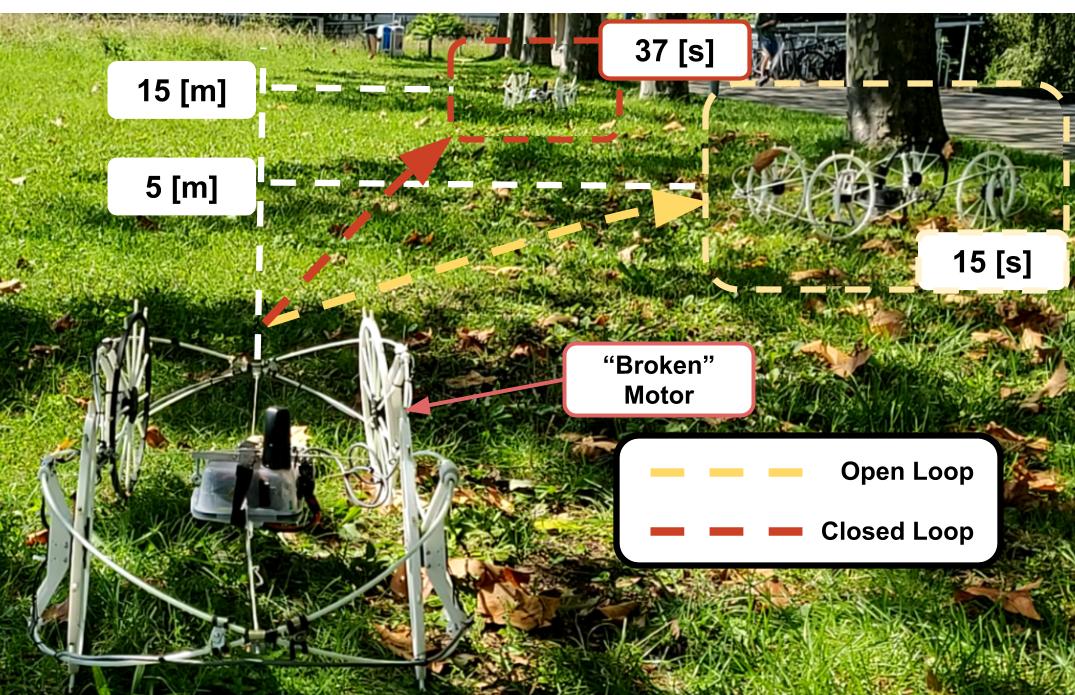}
    \vspace{-4mm}
    \caption{Travel distance and time before going significantly off-course of an outdoor driving experiment with simulated malfunctioning front-right motor.}
    \label{fig:ghost_robots}
    \vspace{-4mm}
\end{figure}

To validate our system further, we conducted a similar straight-line experiment in an outdoor environment. The desired velocity for the front-right wheel was once again multiplied by a constant factor of 0.75 to simulate a malfunction. The resulting final positions of the robot after driving until significantly veering off-course, along with their estimated trajectories and distance travel projected onto the starting position direction, are shown in Fig. \ref{fig:ghost_robots}. The results show that runs utilizing the PID controller with our state estimation managed to travel 300\% farther while maintaining a straighter trajectory in the outdoor environment. For he closed-loop scenario, the final position in Fig. \ref{fig:ghost_robots} is due to accumulation of tracking error over longer period of time that can be fixed with a global referencing such as SLAM. Additionally, unlike the motion capture facility in Fig. \ref{fig:top_view}, without the ability to orient the front wheels (akin to a car), rover mode faces difficulties to turn via only the left and right wheel velocity difference in grassy terrain.


\section{Discussion \& Conclusion}
\label{sec:discussion}

To fully utilize compliant, morphologically adaptive robots, we must exploit their autonomy and control capabilities. However, the key component of any controller is state estimation. Estimating the state of a compliant, morphologically adaptive robot is challenging due to the failure of the rigid-body assumption, compliance with external disturbances, and constant changes in morphology. In this paper, we propose a state estimation algorithm that can overcome these challenges and function under highly uncertain and dynamic robot configurations. We demonstrate that our state estimation framework improves the robot's controllability and increases the system's robustness to sensor noise and mechanical failures.

Our method is capable predicting the shape and velocities of a robot such as GOAT, and the estimation is accurate in domains that we have trained on.  However, as can be seen in Fig. \ref{fig:morph_timeseries}, there is increased reconstruction error in transient morphing states of the robot.
To overcome some of the challenges with unseen data, we would need to record truly dynamic data in a larger motion capture facility such as rolling down a hill and swimming.
Another limitation of this method is the syncing between robot and motion capture data, as well as ensuring that the motion capture quality is high enough for training.
While we have managed to collect enough data, often one minute of training data required up to 10 minutes of post processing to ensure temporal and spatial consistency.
Another venue to train estimators is to augment the data with rigid-body simulations of spheres and driving.

For future work, we plan to build upon this framework to develop a fully autonomous ecosystem. 
This would include a motion controller, a high-level morphology selector, and a high-level navigation module, enabling deployment in challenging outdoor terrain for tasks that rigid-body robots are incapable of tackling.


\section{Acknowledgments}
\label{sec:acknowledgments}
We would like to thank the SYCAMORE Lab and Professor Kamgarpour for lending us the motion tracking facility and providing technical support on the use of the equipment to make this research possible.

\bibliographystyle{IEEEtran}
\bibliography{references}

\end{document}